\DocumentMetadata{uncompress}
\documentclass[letterpaper, 10 pt, conference]{ieeeconf}  

\IEEEoverridecommandlockouts                              

\usepackage{url}
\usepackage{threeparttable}
\usepackage{graphicx}
\usepackage{multirow}
\usepackage{array}
\usepackage{amssymb}
\usepackage{bbding}
\usepackage{booktabs}   
\usepackage{diagbox}
\usepackage{tabularx} 
\usepackage{ragged2e} 
\usepackage{amsmath}
\usepackage{xcolor} 
\usepackage{colortbl}
\usepackage{subcaption}
\usepackage{pifont}

\usepackage{caption}
\usepackage{makecell} 

\title{\LARGE \bf
Seed2Scale: A Self-Evolving Data Engine with Parallel Worlds Expansion for Scalable Robot Learning
}

\author{
    Cong Tai \dag, \ 
    Zhaoyu Zheng \dag, \ 
    Haixu Long \dag, \ 
    Hansheng Wu \dag, \ 
    Jialiang Wang, \
    Zhengbin Long, \ \\
    Haodong Xiang, \
    Shuyuan Wang, \
    Rong Shi, \ 
    Zhuo Cui, \
    Shizhuang Zhang, \ \\ 
    He Wang, \
    Biao Liu, \ 
    Zhenzhe Sun, \ 
    Tao Shen* 
\thanks{\dag\ Equal contribution }
\thanks{* Corresponding author. Emails: shen.tao5@zte.com.cn}
\thanks{All authors are with ZTE Corporation, China.}
}

\let\oldtwocolumn\twocolumn
\renewcommand\twocolumn[1][]{%
    \oldtwocolumn[{#1}{
    \begin{center}
        \vspace{-1.2cm} 
        \includegraphics[width=\textwidth]{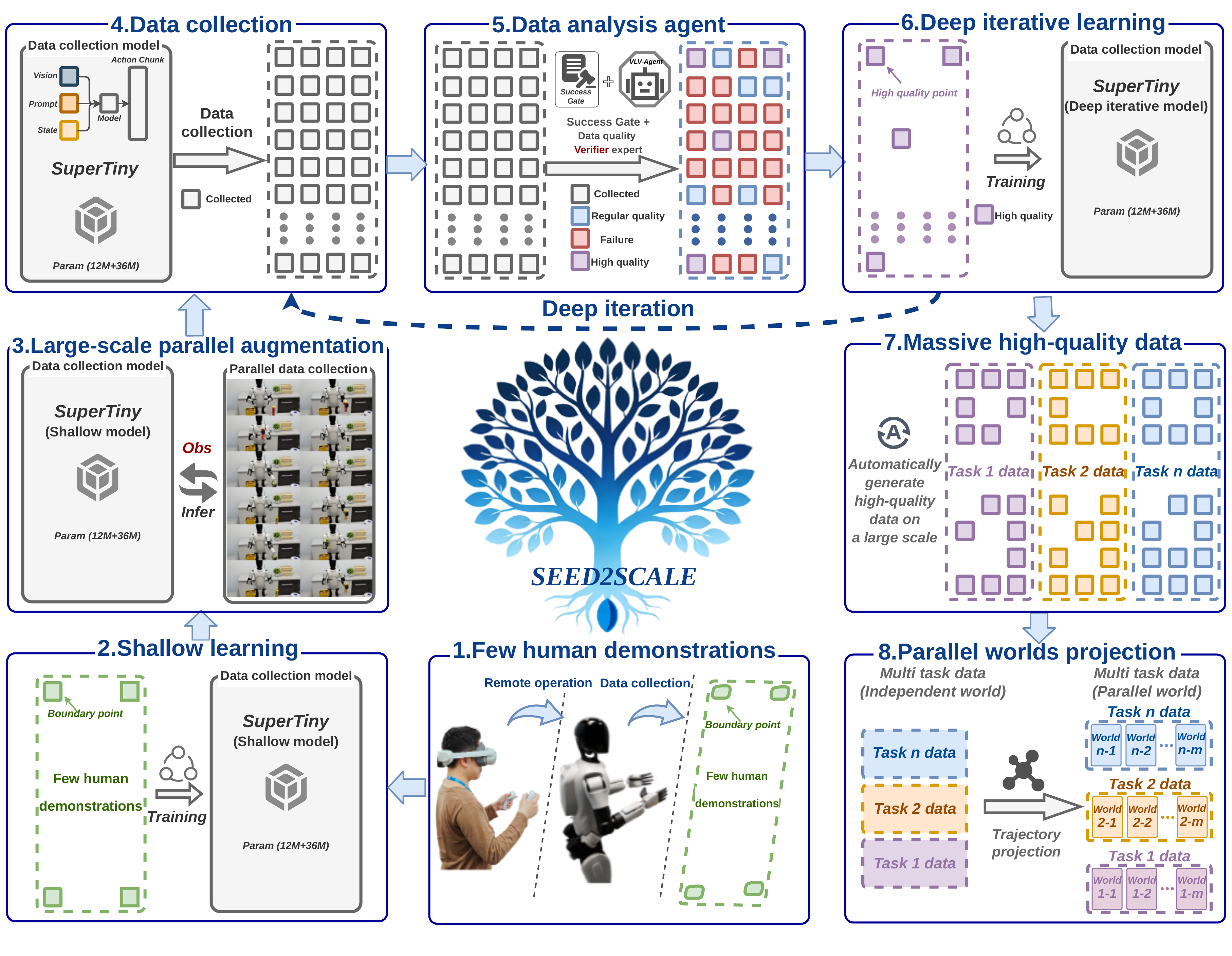}
        \captionof{figure}{The Seed2Scale self-evolving data engine. Starting with as few as four human demonstrations, the framework utilizes a lightweight SuperTiny VLA model, large-scale parallel data collection. A pretrained Vision-Language Model acts as a Verifier, autonomously scoring and filtering the raw trajectories to prevent model collapse. In the next stage, a parallel-world projection mechanism is introduced to achieve visual generalization on the basis of the position generalization accomplished through self-evolution, thereby yielding large-scale, high-quality, and diverse datasets for subsequent model training.}
        \label{fig:teaser}
    \end{center}
    }]
}

\begin{document}
\maketitle

\thispagestyle{empty}
\pagestyle{empty}




\begin{abstract}

Existing data generation methods for robot learning suffer from limited exploration, embodiment gaps, low signal-to-noise ratios, and domain shifts, leading to performance degradation during self-iteration and poor generalization to unseen scenes. 
To address these challenges, we propose Seed2Scale, a self-evolving data engine with parallel worlds expansion. 
Starting with as few as four seed demonstrations, Seed2Scale first executes a self-evolution stage driven by a heterogeneous synergy of ``small-model collection, large-model evaluation, and target-model learning". 
Specifically, the lightweight Vision-Language-Action (VLA) model, SuperTiny, serves as a dedicated data collector for robust exploration. 
Concurrently, a pretrained Vision-Language Model (VLM) functions as a verifier to autonomously score and filter trajectories, supporting stable self-evolution in the evaluated tasks without performance collapse. 
Furthermore, Seed2Scale introduces a parallel worlds stage, projecting self-evolved trajectories into different environments of the same task to generate more diverse data and enhance adaptability to unseen scenes, including real-world environments. 
Experimental results demonstrate that Seed2Scale exhibits significant scaling potential: as iterations progress, the success rate of the target model shows a consistent upward trend, significantly outperforming the seed baseline. Notably, Seed2Scale achieves a remarkable 75.38\% success rate in zero-shot real-world evaluations, where baseline methods fail completely (0\%). 
Project page: https://terminators2025.github.io/Seed2Scale.github.io

\end{abstract}

\section{INTRODUCTION}

The emergence of Vision-Language-Action (VLA) models \cite{shukor2025smolvla, kim2025openvla, chi2023diffusion, intelligence2504pi0, nvidia2025gr00tn16, figure_helix} is accelerating the evolution of Embodied AI, integrating perception, reasoning, and control into a unified neural framework. However, the performance of these models remains strictly contingent upon large-scale, high-quality expert demonstration trajectories \cite{collaboration2023open}. The onerous requirement for manual demonstrations results in a critical ``data scarcity" bottleneck, fundamentally obstructing the scaling of robot learning.

Despite preliminary explorations into automated data generation, existing methods struggle to simultaneously achieve massive scale, high quality, and robust scene generalization. 
Data augmentation-based methods \cite{mandlekar2023mimicgen, jiang2025dexmimicgen, pomponi2025dynamimicgen, ameperosa2025rocoda} can expand expert trajectories through geometric perturbations. However, lacking active exploration, they are restricted to minor perturbations of human demonstrations, unable to generate novel action logic. 
Knowledge transfer from internet videos \cite{ye2024latent, wang2025unified} leverages massive visual data, but due to the inherent ``embodiment gap", it is difficult to precisely translate actions observed in videos into executable commands for physical robots. 
Critically, automated data collection suffers from an extremely low signal-to-noise ratio (SNR). Without effective quality evaluation, unsuccessful trials contaminate the training data, leading to cumulative performance degradation. 
Furthermore, even if spatial diversity is enriched within the source scene, transferring policies to unseen scenes, including real-world environments, remains challenging due to significant domain gaps.

To address these challenges, we propose Seed2Scale, a self-evolving data engine with parallel worlds expansion, designed to alleviate the data scarcity challenge in robot learning. Seed2Scale operates through a dual-stage pipeline: a primary self-evolution stage followed by a parallel worlds projection stage.
In the self-evolution stage, Seed2Scale establishes a heterogeneous synergy architecture of ``small-model collection, large-model evaluation, and target-model learning" to capitalize on the asymmetric advantages of models at different scales. Specifically, we first develop SuperTiny, a lightweight VLA data collector. Leveraging a streamlined query-based architecture and action chunking, SuperTiny achieves robust exploration with as few as four seed trajectories, avoiding the overfitting risks of large models in low-resource settings. Subsequently, the collector generates large-scale trajectories through parallelized rollouts and self-iteration. A pretrained Vision-Language Model \cite{bai2025qwen3, yang2025qwen3} is used as an automated verifier  to provide quality scores for filtering candidate trajectories during self-evolution. This mechanism averts model collapse from low-quality data, ensuring robust self-evolution.
While the self-evolution stage primarily provides spatial generalization within the source scene, the parallel worlds stage projects self-evolved trajectories across different environments of the same task. By synthesizing diverse trajectories, Seed2Scale adapts the policy to unseen domains and real-world deployment.

To evaluate the effectiveness of the proposed method, we conducted extensive experiments across several representative embodied tasks \cite{tai2025realmirror} in both simulation and real-world environments. The experimental results demonstrate the superior performance and scaling potential of Seed2Scale: as the number of iterations increases, the success rate of the target model exhibits a robust consistent trend, improving significantly over the seed baseline, coupled with substantial success in zero-shot real-world evaluations where baseline methods fail completely. 

In summary, the contributions of this work are fourfold:
\begin{itemize}
\item \textbf{Cost-Efficient Self-Evolving Engine:} Enables large-scale data generation from as few as four initial human demonstrations, significantly mitigating the reliance on manual data acquisition in robot learning.
\item \textbf{VLM-Guided Data Curation Pipeline:} Employs a lightweight VLA model, SuperTiny, for robust data collection and a pretrained VLM as a verifier to filter low-quality trajectories, effectively preventing performance degradation during self-iteration.
\item \textbf{Parallel Worlds Expansion Paradigm:} Projects evolved trajectories into diverse environments to improve policy performance on the reported unseen-scene evaluations.
\item \textbf{Experimental Validation and Scaling:} Demonstrates robust performance scaling of the target model across iterations, significantly outperforming the seed baseline, and achieves a 75.38\% success rate in the reported zero-shot real-world pick-and-place evaluation, without real-world fine-tuning.
\end{itemize}

\section{RELATED WORK}



\subsection{Vision-Language-Action}

Vision-Language-Action (VLA) models integrate visual-language understanding with robot action prediction. OpenVLA \cite{kim2024openvla} autoregressively predicts discretized action tokens and demonstrates cross-task transfer. Recent work combines large-scale pretraining with continuous action generation: $\pi_{0.5}$ \cite{intelligence2504pi0} uses a flow-matching action expert during post-training for cross-task and cross-embodiment generalization, while NVIDIA's GR00T N1.6 \cite{nvidia2025gr00tn16} and Figure AI's Helix \cite{figure_helix} target scalable humanoid control. SmolVLA \cite{shukor2025smolvla} uses a compact flow-matching architecture for resource-efficient, real-time deployment.

\subsection{Automated Data Generation and Augmentation}

To mitigate the prohibitive costs of human demonstrations, demonstration-based offline augmentation and retargeting expand expert datasets via geometric perturbations and inverse-kinematics transformations \cite{mandlekar2023mimicgen, ameperosa2025rocoda, jiang2025dexmimicgen, pomponi2025dynamimicgen}. Recent methods diversify visual worlds through 3D scene editing and open-world synthesis while preserving executable actions \cite{zhao2026real2edit2real}. Validated visual and spatial perturbations correlate with real-world VLA generalization, motivating our parallel-world rollout augmentation \cite{sedlacek2026realm}. For contact-rich skills, physics-constrained synthesis combines VLM guidance with simulation to enforce contact-dynamics feasibility \cite{jing2025humanoidgen}.
To shatter real-world data barriers, video-to-robot weakly supervised learning leverages massive internet videos via explicit cross-embodiment mapping \cite{bahl2022human, dessalene2025embodiswap, liu2025immimic}, physics-grounded reconstruction \cite{hsieh2025dexman}, and scaling validations \cite{zheng2026egoscale}, and virtual-real simulation platforms for robotic task development and evaluation \cite{zheng2025demonstrating}, while increasingly evolving toward implicit generative priors like discrete latent actions \cite{ye2024latent}, inverse-dynamics world models \cite{ye2026world}, and video pretraining \cite{wu2023unleashing}.



\subsection{Vision-Language Models}

VLMs have shown strong capabilities in visual understanding, language grounding, and video comprehension, making them promising tools for embodied AI. Recent models \cite{bai2025qwen3, chen2024internvl, hurst2024gpt} further advance complex scene interpretation, long-horizon video understanding, and fine-grained spatial perception.These capabilities make VLMs well-suited for evaluation-centric roles in embodied AI, such as outcome assessment and trajectory quality evaluation, which motivates their use as automated verifiers in our pipeline. 


\begin{figure*}[t]
    \centering
    \includegraphics[width=1.0\linewidth]{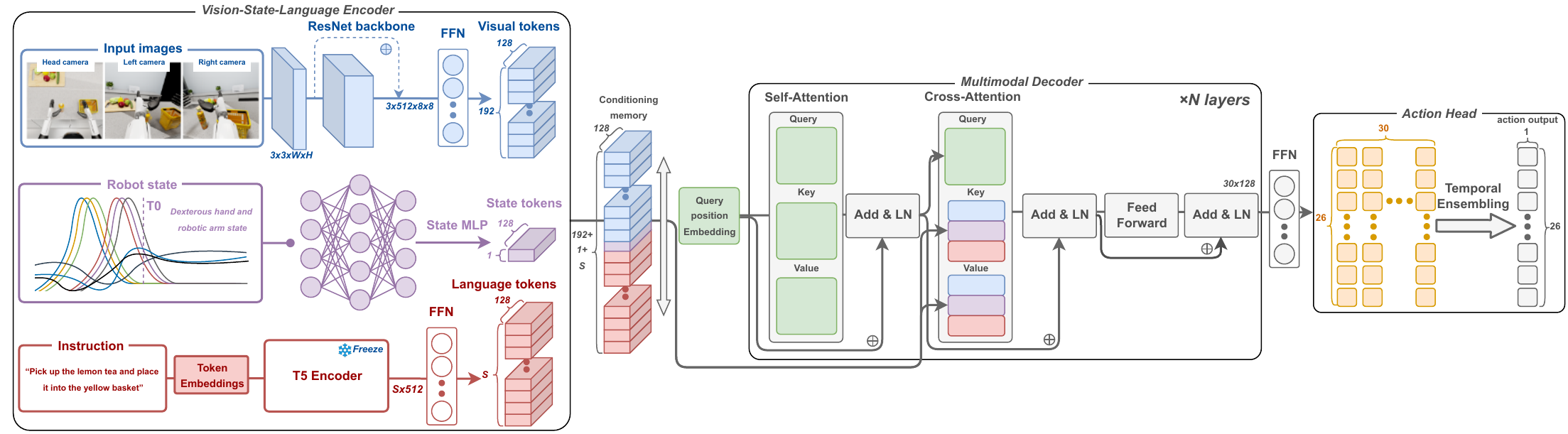}
    \caption{Architecture of the SuperTiny VLA. The model integrates vision (ResNet), language (T5), and robot state encodings into a unified conditioning memory $\mathcal{M}_t$, which is processed by a lightweight Transformer decoder via cross-attention to predict action chunks. Temporal ensembling (Eq. \ref{eq:temporal_ensembling}) ensures smooth and consistent control.}
    \label{fig:supertinyvla_arch}
    \vspace{-4mm}
\end{figure*}
\section{Seed2Scale}


The synergy of Seed2Scale comprises three specialized model roles and one
trajectory-level augmentation stage:

\begin{itemize}
    \item \textbf{Small-Scale Collector ($\pi_{small}$):} A lightweight VLA
    enabling high-throughput parallel rollouts and seed-data bootstrapping.

    \item \textbf{Large-Scale Verifier ($\Phi_{VLV}$):} A frozen Qwen3-VL
    \cite{bai2025qwen3, yang2025qwen3} that scores and filters candidate
    trajectories without human intervention.

    \item \textbf{Parallel Worlds:} A trajectory-conditioned stage that
    projects final self-evolved trajectories into Parallel Worlds,
    retargets their actions, and re-collects quality-filtered rollouts for
    final generalization.

    \item \textbf{Target Model ($\pi_{target}$):} SmolVLA
    \cite{shukor2025smolvla}, trained on the curated dataset
    $\mathcal{D}_{final}$ to distill verified visuomotor behavior.

\end{itemize}

By orchestrating these heterogeneous components, the Seed2Scale engine transforms a sparse set of 4 human demonstrations into a continuous flow of high-quality, verified training data, thereby bridging the gap between minimal expert input and generalized embodied intelligence. The workflow of the proposed framework, from seed bootstrapping to target model scaling, is illustrated in Fig. \ref{fig:teaser}. 

\subsection{SuperTiny: Small-Scale VLA Collector}
\label{subsec:supertiny}

For high-throughput data collection, we develop \textit{SuperTiny}, a lightweight VLA model for efficient inference and low-resource bootstrapping. As shown in Fig.~\ref{fig:supertinyvla_arch}, a heterogeneous encoding strategy unifies vision, language, and robot state. A ResNet-18 backbone extracts visual features, projected via a 1×1 convolution. A pretrained T5-Small encodes language instructions. A compact MLP embeds robot state. All modalities are concatenated into a conditioning memory:

\begin{equation}
    \mathcal{M}_t = [z_{\mathrm{vis}}; z_{\mathrm{lang}}; z_{\mathrm{state}}]
\end{equation}


A lightweight Transformer decoder processes learnable query embeddings $Q$ via cross-attention over $\mathcal{M}_t$, and the decoder output is then projected through an action head to predict action chunks:
\begin{equation}
    A_t = \textit{ActionHead}(\textit{Decoder}(Q, \mathcal{M}_t)) \in \mathbb{R}^{K \times d_a}
\end{equation}
where $K$ is the chunk size, and $d_a$ is the action dimensionality.

To ensure smooth control during rollouts, we apply exponential temporal ensembling over overlapping action chunks:
\begin{equation}
\label{eq:temporal_ensembling}
    a_t = \sum_{k=0}^{K-1} w_k \cdot A_{t-k}[k], \quad w_k = \frac{\exp(-\lambda k)}{\sum_{j=0}^{K-1} \exp(-\lambda j)}
\end{equation}
where $A_{t-k}[k]$ is the $k$-th action in the chunk predicted at time $t{-}k$, and $\lambda$ controls the decay rate.


\subsection{VLM-as-a-Verifier}
\label{subsec:vlv}


The primary challenge in automated data collection is the low  SNR of raw trajectories. Without effective quality evaluation, failed explorations contaminate the training data and trigger performance degradation during self-iteration. To address this, we adopt a two-stage verification pipeline. First, a rule-based success gate filters out failed trajectories, ensuring by design that they do not enter the training set. The remaining trajectories are then scored for execution quality by a frozen, pretrained VLM (Qwen3-VL~\cite{bai2025qwen3, yang2025qwen3}), which serves as a Vision-Language Verifier (VLV) to score and filter generated trajectories.

For each successful trajectory $\tau \in \mathcal{T}_{succ}$, the VLV receives a multimodal prompt consisting of: (1) the task instruction $g$, (2) the rollout video $V_{\tau}$ of the current attempt, and (3) a reference video $V_{ref}$ demonstrating successful execution from the seed data. The verifier outputs a quality score:
\begin{equation}
    S_{\tau} = \Phi_{VLV}(V_{\tau}, V_{ref}, g) \in [0, 10]
\end{equation}
\vspace{2mm}
where higher scores indicate better task completion quality.

Based on these scores, only trajectories exceeding a quality threshold $\gamma$ are admitted into the curated dataset $\mathcal{D}_{silver}$. The threshold $\gamma$ was fixed a priori according to a manually designed 0–10 quality rubric. It was not optimized using the downstream test performance or evaluation scenes. This selective filtering restricts target-model training to trajectories that satisfy the VLV’s predefined quality criterion.

\subsection{Seed2Scale Bootstrapping}
\label{subsec:bootstrapping}


To minimize reliance on expensive human demonstrations, For each evaluated task, Seed2Scale is initialized with four expert trajectories $\mathcal{D}_{seed}$ collected from four boundary or corner configurations, which correspond to the four corner positions of the tabletop workspace. Seed2Scale comprises two main stages: (1) self-evolution, a recursive data expansion loop that broadens the exploration frontier beyond the seed distribution for spatial generalization; (2) Parallel Worlds, which collect data via action projection in diverse environments for trajectory-based visual generalization, facilitating VLA transfer to unseen and real-world scenarios.

\textbf{Self-evolution:}

At iteration $i$, the lightweight collector $\pi_{small}^{(i)}$ is trained on dataset $\mathcal{D}^{(i)}$ (initialized as $\mathcal{D}^{(0)} = \mathcal{D}_{seed}$), then deployed in $N_{env}$ parallel environments to generate raw trajectories:
\begin{equation}
    \mathcal{T}_{raw}^{(i)} = \{ \tau_j \sim \pi_{small}^{(i)}(\cdot | I, g, s) \}_{j=1}^{N_{rollout}}
\end{equation}
where each trajectory $\tau_j = \{(I_t, s_t, a_t, g)\}_{t=0}^{T_j}$ contains visual observations, robot states, actions, and task instructions.

The VLV scores each trajectory $\tau_j$ (Sec.~\ref{subsec:vlv}). Only high-quality samples exceeding threshold $\gamma$ are retained:
\begin{equation}
    \mathcal{D}_{silver}^{(i)} = \{ \tau_j \in \mathcal{T}_{succ}^{(i)} \mid S_{\tau_j} \geq \gamma \}
\end{equation}
The dataset is then augmented for the next iteration:
\begin{equation}
    \mathcal{D}^{(i+1)} = \mathcal{D}^{(i)} \cup \mathcal{D}_{silver}^{(i)}
\end{equation}


At this stage, we call this variant \textbf{Seed2Scale$^{*}$}, which includes only the Self-evolution stage and provides spatially generalized data in the source world.

\textbf{Parallel Worlds:}

After self-evolution, we denote the final source-world dataset by $\mathcal{D}_{{se}}=\mathcal{D}^{(I)}$, where $I$ is the final iteration in the self-evolution stage. To improve visual generalization, each
source trajectory is projected in multiple target worlds with different
visual contexts. A target world is specified as:
\begin{equation}
\mathcal{C}_m =
\left(
\mathcal{A}_m,\mathcal{P}_F,\mathcal{B}_m,\mathcal{K}_m
\right)
\label{eq:pw_scene}
\end{equation}
where $\mathcal{A}_m$, $\mathcal{P}_F$, $\mathcal{B}_m$, and
$\mathcal{K}_m$ denote the visual assets, source-aligned object poses,
background and appearance configuration, and camera configuration,
respectively. The task-relevant objects in all target worlds are
placed at the same poses as in the source world. 

Given a source trajectory
$\tau_F=\{(I_t^F,s_t^F,a_t,g)\}_{t=0}^{T}$, we project its control
trajectory into the target world under source-aligned task geometry.
This design follows the general scene-aware trajectory transformation
paradigm, in which a demonstrated trajectory is transferred to a
target scene before execution~\cite{heppert2024ditto}:
\begin{equation}
\mathcal{M}_m(\tau_F)
=
\begin{cases}
\tau_F, & \tau_F\in\mathcal{F}(\mathcal{C}_m),\\
\varnothing, & \text{otherwise}.
\end{cases}
\label{eq:pw_mapping}
\end{equation}
where $\mathcal{M}_m$ denotes the source-to-target projection operator
and $\mathcal{F}(\mathcal{C}_m)$ denotes the target-world feasibility
check. Specifically, the check verifies workspace validity, kinematic
validity, collision avoidance, contact consistency, and trajectory
smoothness. If the projected trajectory is feasible, its original
control sequence is executed in the target world to collect new
observations and robot states, and the resulting trajectory is filtered
by the VLV:
\begin{equation}
\mathcal{D}_{\mathrm{final}}
=
\mathcal{D}_{\mathrm{se}}
\cup
\bigcup_{\tau_F\in\mathcal{D}_{\mathrm{se}}}
\bigcup_{m=1}^{M}
\mathcal{Q}_{\gamma}
\left[
\operatorname{Rollout}_{\mathcal{C}_m}
\left(\mathcal{M}_m(\tau_F)\right)
\right]
\label{eq:pw_dataset}
\end{equation}

Throughout the remainder of the paper, Seed2Scale$^{*}$ denotes the Self-evolution-only variant, the Parallel Worlds stage contributes to trajectory-based visual generalization, and Seed2Scale denotes their complete sequential combination. By leveraging high inference throughput, the engine, jointly driven by the self-evolution and Parallel Worlds, synthesizes thousands of diverse trajectories without manual intervention, overcoming the traditional data scarcity limitation in Embodied AI.


\subsection{Target Model Training}
\label{subsec:target_training}


The target model $\pi_{target}$ (SmolVLA~\cite{shukor2025smolvla}) is trained with Conditional Flow Matching to generate structured action sequences through iterative denoising, enabling robust policy learning from $\mathcal{D}_{final}$.

Given visual and linguistic features $o_t$ extracted by the VLM backbone, the model learns a vector field $v_\theta$ that maps noisy actions $A^\sigma_t = \sigma A_t + (1-\sigma)\epsilon$ to the target vector field $u_t = A_t - \epsilon$, where $\epsilon \sim \mathcal{N}(0, I)$ is Gaussian noise and $\sigma \in [0, 1]$ is the flow step. The training objective is:
\begin{equation}
    \mathcal{L}_{FM}(\theta) = \mathbb{E}_{\sigma, \epsilon, A_t, o_t} \left[ \| v_\theta(A^\sigma_t, o_t) - u_t \|_2^2 \right]
\end{equation}

To ensure temporally coherent action chunks $A_t = (a_t, \dots, a_{t+K})$, SmolVLA employs an Action Expert architecture that interleaves Cross-Attention (CA) with Self-Attention (SA):
\begin{equation}
    \textit{ActionExpert}(o_t) = \textit{SA}(\textit{CA}(Q_{actions}, K_{o_t}, V_{o_t}))
\end{equation}
The CA layers attend to vision-language context $o_t$, while the SA layers model temporal dependencies within the action sequence. This design implicitly enforces smoothness without requiring explicit regularization terms, resulting in stable closed-loop control during deployment.

\section{Experiments}
\label{sec:experiments}


We design experiments to answer the following questions:
(1) Can Seed2Scale significantly improve VLA policy performance with minimal seed data?
(2) How does our work perform in real-world scenarios?
(3) Does the self-evolution mechanism exhibit consistent scaling performance across iterations?
(4) How does the self-evolution method compare with existing data augmentation methods?
(5) Can the lightweight SuperTiny collector achieve competitive performance while enabling efficient data generation?
(6) Does the VLV effectively serve as a quality gatekeeper?

\subsection{Experimental Setup}
\label{subsec:setup}

\begin{figure}[!ht]
    \vspace{-4mm}
    \centering
    \includegraphics[width=0.8\linewidth]{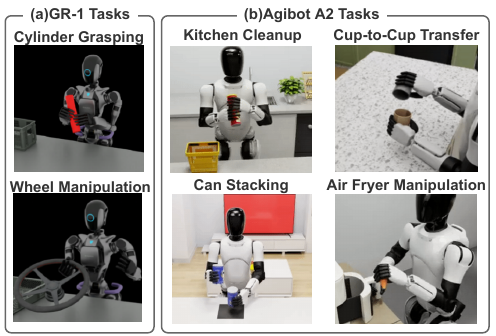}
    \caption{(a) GR-1 robot tasks used for comparison with MimicGen and Seed2Scale$^{*}$. (b) Agibot A2 robot tasks for Seed2Scale evaluation.}
    \label{fig:4task}
    \vspace{-3mm}
\end{figure}

We evaluate Seed2Scale on the manipulation suite in
Fig.~\ref{fig:4task}, which comprises four Agibot A2 tasks on the
RealMirror~\cite{tai2025realmirror} platform and two GR-1 tasks against the
MimicGen~\cite{mandlekar2023mimicgen} baseline. 

\begin{table*}[!t]
\centering
\footnotesize
\caption{Success rate (\%) of the target model on the four Agibot A2
tasks, evaluated on Seen and Unseen scenes over 396  independent trials, with best results per task in \textbf{bold}. For each variant, a single SmolVLA is trained on data combined from all four tasks and evaluated separately on each task.}
\label{tab:main_results}
\begin{tabular}{l ccc ccc}
\toprule
& \multicolumn{3}{c}{Evaluated on Seen Scenes} & \multicolumn{3}{c}{Evaluated on Unseen Scenes} \\
\cmidrule(lr){2-4} \cmidrule(lr){5-7}
\textbf{Task} & \textbf{SeedOnly} & \textbf{Seed2Scale$^{*}$} & \textbf{Seed2Scale} & \textbf{SeedOnly} & \textbf{Seed2Scale$^{*}$} & \textbf{Seed2Scale} \\
\midrule
Kitchen Cleanup        & 24.63\% & \textbf{71.43\%} & 68.50\% &  0.00\% &  0.00\% & \textbf{75.51\%} \\
Cup-to-Cup Transfer    & 23.50\% & \textbf{64.14\%} & 60.61\% &  0.00\% &  0.00\% & \textbf{61.62\%} \\
Can Stacking           &  7.50\% & 65.90\% & \textbf{67.70\%} &  0.00\% &  0.00\% & \textbf{69.19\%} \\
Air Fryer Manipulation & 33.08\% & \textbf{72.82\%} & 71.21\% &  0.00\% &  0.00\% & \textbf{55.05\%} \\
\midrule
Average                & 22.18\% & \textbf{68.57\%} & 67.01\% &  0.00\% &  0.00\% & \textbf{65.34\%} \\
\bottomrule
\end{tabular}
\vspace{-5mm}
\end{table*}



\begin{figure}[t]
    \centering
    \includegraphics[width=0.95\linewidth]{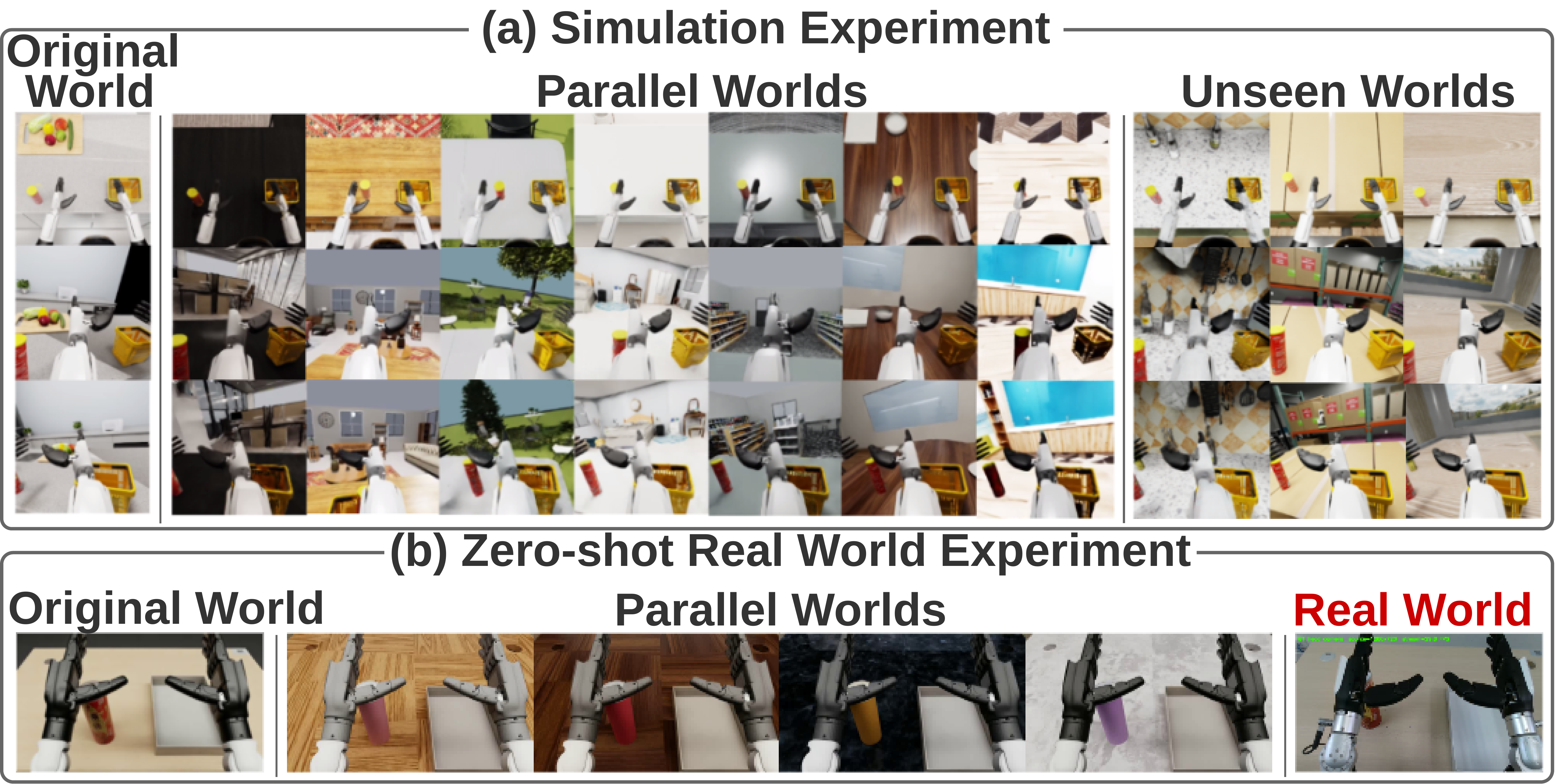}
    \caption{(a) The original world, the parallel worlds projected from it, and the unseen worlds. (b) The worlds in the zero-shot real-world experiment.}
    \label{fig:mutil_view}
    \vspace{-5mm}
\end{figure}

The Parallel Worlds consist of 10 distinct scenes for the A2
task suite, with variation in object instances, spatial arrangement, and
background context, as illustrated in Fig.~\ref{fig:mutil_view}(a), some of
which are constructed with 3D Gaussian Splatting (3DGS). Seen denotes the
10 scenes used for data generation, whereas Unseen denotes three additional
disjoint scenes reserved for evaluation and excluded from training.

Each task is initialized with the original four seed demonstrations collected
by teleoperation. In Parallel Worlds, the self-evolved trajectories are paired with
regenerated scene-conditioned observations in the additional scenes.
We use Success Rate (\%) as the primary
metric for model evaluation. Additional metrics are reported where applicable.

The SuperTiny VLA collector contains 48M parameters with action chunk size
$K=30$, and the VLV is implemented using Qwen3-VL-32B~\cite{bai2025qwen3}.
The target model SmolVLA is trained for 100,000 steps using the AdamW optimizer
($lr=3\times10^{-4}$, batch size 16).

\subsection{Simulation Performance Improvement and Parallel Worlds Generalization}
\label{subsec:main_results}

We first investigate whether Seed2Scale can significantly improve downstream policy performance from minimal seed data, and whether the Parallel Worlds expansion extends this generalization to unseen scenes. Table~\ref{tab:main_results} reports the success rates on the four Agibot A2 tasks, evaluated separately on Seen and Unseen scenes.

The results reveal two distinct phases of capability acquisition. First, the self-evolution pipeline alone (\textbf{Seed2Scale$^{*}$}) boosts the average success rate on Seen scenes from $22.18\%$ (\textbf{SeedOnly}) to $68.57\%$, validating the scaling benefit of self-evolution. 
Notably, Can Stacking exhibits the most significant improvement ($7.50\% \rightarrow 65.90\%$), as the original 4-seed dataset provides
insufficient coverage for the complex spatial alignments and
stable placement required for stacking. 
We also observe that joint training on multiple tasks yields higher performance than single task training (Fig. \ref{fig:scaling_curve}), demonstrating beneficial cross-task
knowledge learning.

Second, both SeedOnly and Seed2Scale$^{*}$ fail completely on Unseen scenes ($0\%$ success rate), confirming that trajectory scaling confined to a single source world cannot bridge the visual domain gap. Incorporating Parallel Worlds (\textbf{Seed2Scale}) resolves this limitation: it preserves the high performance on Seen scenes ($67.01\%$) while dramatically elevating the Unseen success rate from $0\%$ to an average of $65.34\%$. This isolates the contribution
of scene diversity from that of additional action supervision.

These results support a scene-coverage-driven explanation of generalization: by re-projecting self-evolved trajectories across heterogeneous environments, Seed2Scale decorrelates the policy from source-scene appearance. This yields robust competence in unseen domains without compromising in-distribution accuracy.

\subsection{Zero-Shot Generalization in Real-World Scenes}
\label{subsec:real_robot}

To examine whether the scene diversity introduced by Parallel Worlds can improve transfer to a physical robot, we deploy the policies trained exclusively in simulation. Crucially, no real-world data collection or fine-tuning is performed. The evaluation adopts a typical \textit{Pick-and-Place} task, with 7 scenes for training data and the real
scene for evaluation, as illustrated in Fig.~\ref{fig:mutil_view}(b).

\begin{table}[t]
\centering
\caption{Zero-shot success rate on the Pick-and-Place task,
evaluated over 65 independent trials in an unseen real-world scene. Seed2Scale ($n$ scenes) denotes a configuration trained with parallel-world data from $n$ of the 7 training scenes.}
\label{tab:real_robot}
\begin{tabular}{l c c c}
\toprule
\textbf{Configuration} & \textbf{Episodes} & \textbf{Frames}& \textbf{Success Rate} \\
\midrule
\textbf{SeedOnly}                  &4 &720 &0.00\% \\
\textbf{Seed2Scale$^{*}$}          &142 &19,748 &0.00\% \\
\textbf{Seed2Scale (3 scenes)}     &432 &60,039 &27.69\% \\
\textbf{Seed2Scale (5 scenes)}     &720 &100,065 &73.85\% \\
\textbf{Seed2Scale (7 scenes)}     &1,716 &225,217 &\textbf{75.38\%} \\
\bottomrule
\end{tabular}
\vspace{-5mm}
\end{table}

Table~\ref{tab:real_robot} reports the zero-shot success rate of each configuration on the physical robot.  The
 training data expand from 4 episodes and 720 frames for SeedOnly to 1,716 episodes and 225,217 frames for Seed2Scale trained with 7 scenes. In contrast to the baselines, Seed2Scale successfully transfers to the real world, with its success rate increasing monotonically alongside scene diversity: 27.69\% with 3 scenes, 73.85\% with 5 scenes, and 75.38\% with 7 scenes. This striking trend—from 0\% in the single source world to 75.38\% across 7 parallel scenes—demonstrates that Parallel Worlds is the primary driver of zero-shot generalization to real-world environments. The smaller marginal gain from 5 to 7 scenes is consistent with the diminishing returns of additional scene diversity.

\subsection{Scaling Performance of the Self-Evolution Mechanism Across Iterations}
\label{subsec:scaling}




\begin{figure}[ht]
    \centering
    \includegraphics[width=0.7\linewidth]{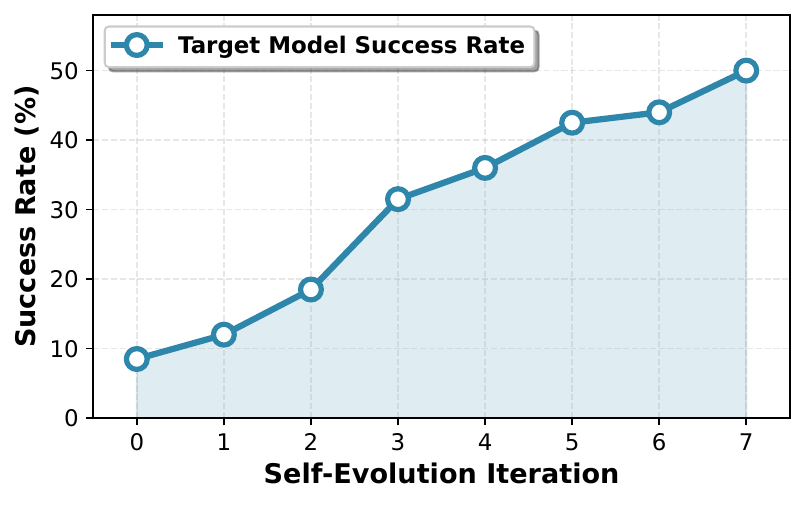}
    \caption{Scaling performance of the target model (SmolVLA) across self-evolution iterations.}
    \label{fig:scaling_curve}
    \vspace{-2mm}
\end{figure}
To investigate the scaling properties of the self-evolution, we track the target model's performance on the most challenging Can Stacking task across 8 self-evolution iterations. Specifically, at each iteration $i$, we use the accumulated Can Stacking data generated by SuperTiny (up to iteration $i$) to train a SmolVLA model from scratch, then evaluate its success rate. This yields 8 independent single-task SmolVLA checkpoints, allowing us to observe how target model performance scales with the self-evolution process.

As shown in Fig. \ref{fig:scaling_curve}, starting from merely 4 seed demonstrations, the success rate follows a consistent upward trend as iterations progress, achieving substantial performance gains through the self-evolution mechanism.

\subsection{Comparison with Existing Data Augmentation Methods}
\label{subsec:mimicgen}


To validate the benefit of self-evolving exploration over conventional augmentation, we compare it with MimicGen\cite{mandlekar2023mimicgen} on GR-1 tasks. Both methods are evaluated under MimicGen's experimental configuration for a fair comparison. As shown in Table \ref{tab:mimicgen_comparison}, self-evolution outperforms MimicGen on both metrics, achieving 86.96\% replay success—a fourfold improvement—and downstream performance gains of 28.75\% on Cylinder Grasping and 58.50\% on Wheel Manipulation. MimicGen’s inverse-kinematics-based trajectories can be kinematically infeasible due to the absence of closed-loop environmental feedback. We further compare human demonstrations, MimicGen data, and self-evolved trajectories using three joint-motion metrics. 
\begin{table}[t]
\centering
\caption{Comparison with MimicGen on GR-1 tasks. We report the policy and replay success rates. Seed2Scale$^{*}$ outperforms MimicGen on both.}
\label{tab:mimicgen_comparison}
\begin{tabular}{llcc}
\toprule
\textbf{Metric} & \textbf{Task} & \textbf{MimicGen} & \textbf{Seed2Scale$^{*}$} \\
\midrule
\multirow{3}{*}{\textbf{Policy Succ.}} 
& Cylinder Grasp & 37.25\% & \textbf{66.00\%} \\
& Wheel Manip. & 34.75\% & \textbf{93.25\%} \\
\cmidrule{2-4}
& \textit{Average} & 36.00\% & \textbf{79.63\%} \\
\midrule
\multirow{3}{*}{\textbf{Replay Succ.}} 
& Cylinder Grasp & 21.00\% & \textbf{86.96\%} \\
& Wheel Manip. & 48.50\% & \textbf{67.86\%} \\
\cmidrule{2-4}
& \textit{Average} & 34.75\% & \textbf{77.41\%} \\
\bottomrule
\end{tabular}
\vspace{-4mm}
\end{table}


\begin{itemize}
    \item \textbf{Total Variation (TV):} measures the cumulative magnitude of trajectory changes; lower values indicate smoother trajectories.

    \item \textbf{Mean Absolute Jerk:} Jerk represents the rate of change of acceleration. It measures motion naturalness.

    \item \textbf{High-Frequency (HF) Ratio:} FFT-based proportion of high-frequency energy relative to total non-DC energy.
\end{itemize}

\begin{table}[h]
    \vspace{-2mm}
    \centering
    \caption{Quantitative comparison of trajectory quality. Lower values indicate better performance. Seed2Scale$^{*}$ produces trajectories with human-like smoothness, significantly outperforming MimicGen.}
    \label{tab:trajectory_quality}
    
    \begin{tabular}{l@{\hspace{3pt}}c@{\hspace{6pt}}c@{\hspace{6pt}}c}
        \toprule
        \textbf{Metric} & \textbf{Expert Demonstration} & \textbf{MimicGen} & \textbf{Seed2Scale$^{*}$} \\
        \midrule
        Total Variation      & 1.32 & 3.68 & 1.34 \\
        Mean Absolute Jerk       & 0.0063 & 0.0261 & 0.0047 \\
        HF Power Ratio (\%)     & 0.22 & 2.07 & 0.30 \\
        \bottomrule
        \end{tabular}
    \vspace{-2mm}
\end{table}

\begin{figure}[t]
    \centering
    \includegraphics[width=0.9\linewidth]{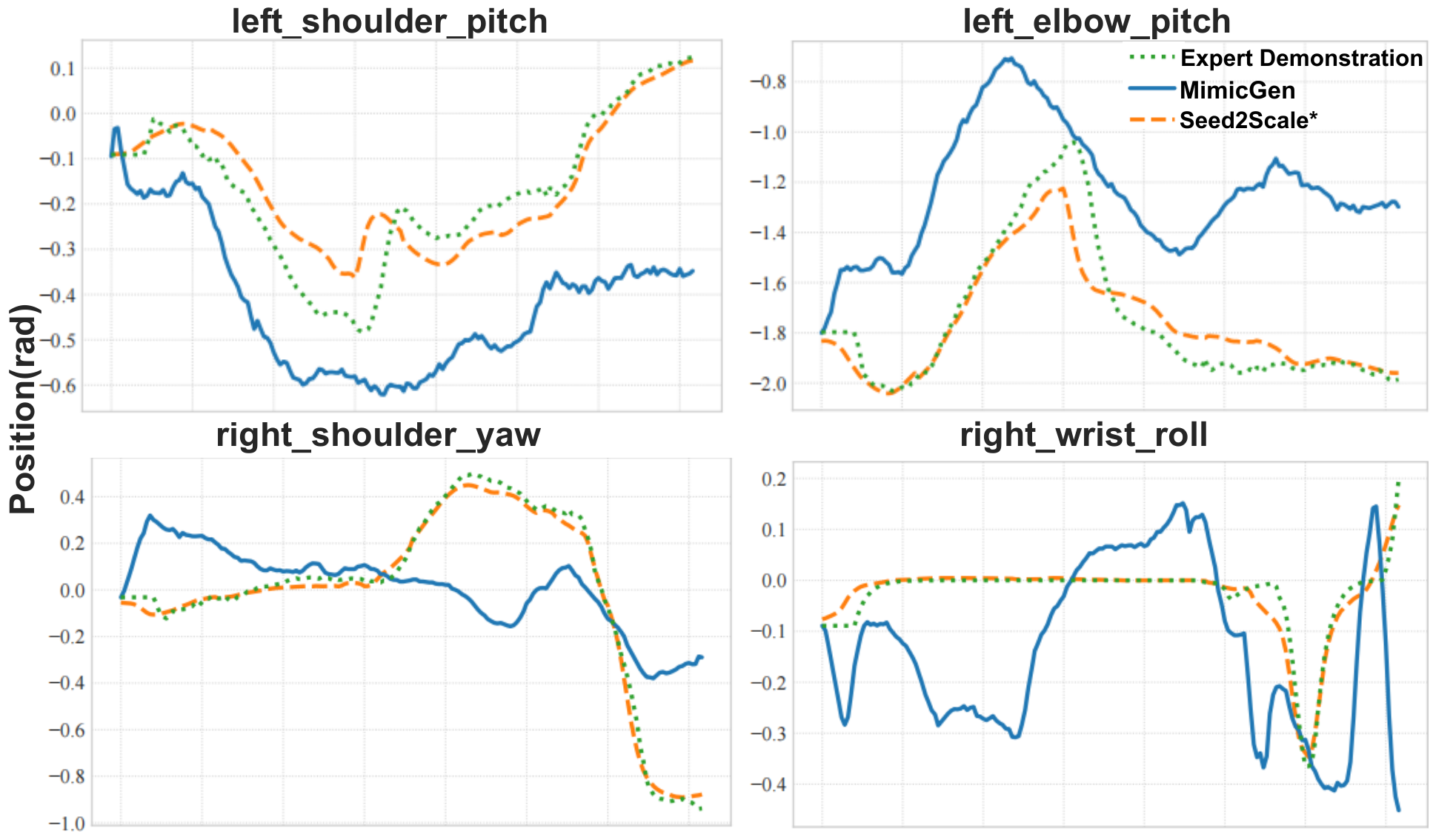}
    \caption{Comparison of robot action curves generated by expert demonstration, MimicGen, and the Seed2Scale$^{*}$.}
    \label{fig:motor_curve}
    \vspace{-5mm}
\end{figure}

The experimental results are shown in Fig.~\ref{fig:motor_curve}. Trajectories generated by the self-evolution method closely match human demonstrations. Unlike MimicGen, which suffers from severe high-frequency jitter due to its IK-based transformations, our method produces smooth and stable motion curves that follow the expert profile.

The quantitative results in Table~\ref{tab:trajectory_quality} further confirm this improvement. MimicGen exhibits nearly $3\times$ higher Total Variation and $5.5\times$ higher Jerk, indicating substantial jitter and physical infeasibility caused by IK-based transformations. In comparison, our method achieves a Total Variation of 1.34 and an HF Ratio of 0.30\%, both close to human demonstrations. Its Jerk is also lower than that of human data, Under these metrics, Seed2Scale* produces smoother joint motions than MimicGen, closely matching expert demonstrations while preserving task behavior.

\subsection{Efficiency and Competitive Performance of the SuperTiny Collector}
\label{subsec:supertiny_eval}


A core design choice in the self-evolution is the use of a lightweight VLA model as the data collector. To validate that SuperTiny achieves competitive task performance, we compare it against two representative VLA architectures: ACT \cite{zhao2023learning} and Diffusion Policy \cite{chi2023diffusion}. 

\begin{table}[t]
    \centering
    \caption{Efficiency comparison of VLA models.}
    \label{tab:efficiency}
    \setlength{\tabcolsep}{6pt}
    \begin{tabular}{lccc}
        \toprule
        \textbf{Model} & \textbf{Params} & \textbf{Inference Time} & \textbf{Frequency} \\
        \midrule
        ACT & 52M & 45.67ms & 21.9 Hz \\
        Diffusion Policy & 265M & 135.83ms & 7.4 Hz \\
        \textbf{SuperTiny} & \textbf{48M} & \textbf{38.08ms} & \textbf{26.3 Hz} \\
        \bottomrule
    \end{tabular}
    \vspace{-3mm}
\end{table}

Table \ref{tab:efficiency} compares computational efficiency. SuperTiny achieves the fastest inference time (38.08ms) with a real-time control frequency of 26.3 Hz---1.2$\times$ faster than ACT and 3.6$\times$ faster than Diffusion Policy. With only 48M parameters, SuperTiny is 5.5$\times$ smaller than Diffusion Policy. This efficiency enables large-scale parallel data collection.

\begin{figure}[t]
    \centering
    \includegraphics[width=0.8\linewidth]{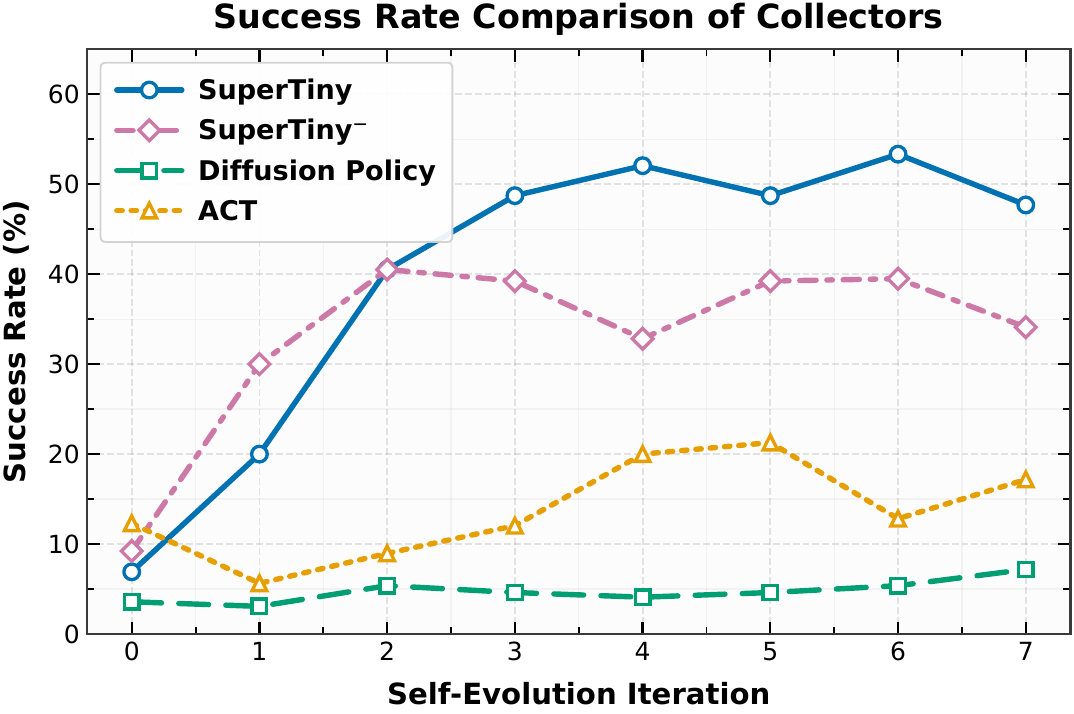}
    \caption{Performance comparison of SuperTiny, ACT, and Diffusion Policy as data collectors across self-evolution iterations. SuperTiny$^-$ denotes the variant without VLV-based quality filtering.}
    \label{fig:collector_comparison}
    \vspace{-3mm}
\end{figure}

Fig. \ref{fig:collector_comparison} compares the scaling performance of SuperTiny, ACT, and Diffusion Policy as data collectors. All models exhibit improved final performance, validating the effectiveness of our data selection and self-evolution mechanism. Notably, SuperTiny converges faster and achieves higher early success rates than the baselines, and reaches the highest final success rate, outperforming ACT and Diffusion Policy.

SuperTiny's combination of real-time efficiency and strong task performance makes it the ideal collector for the self-evolution stage.

\begin{figure}[!ht]
    \centering
    \includegraphics[width=1.0\linewidth]{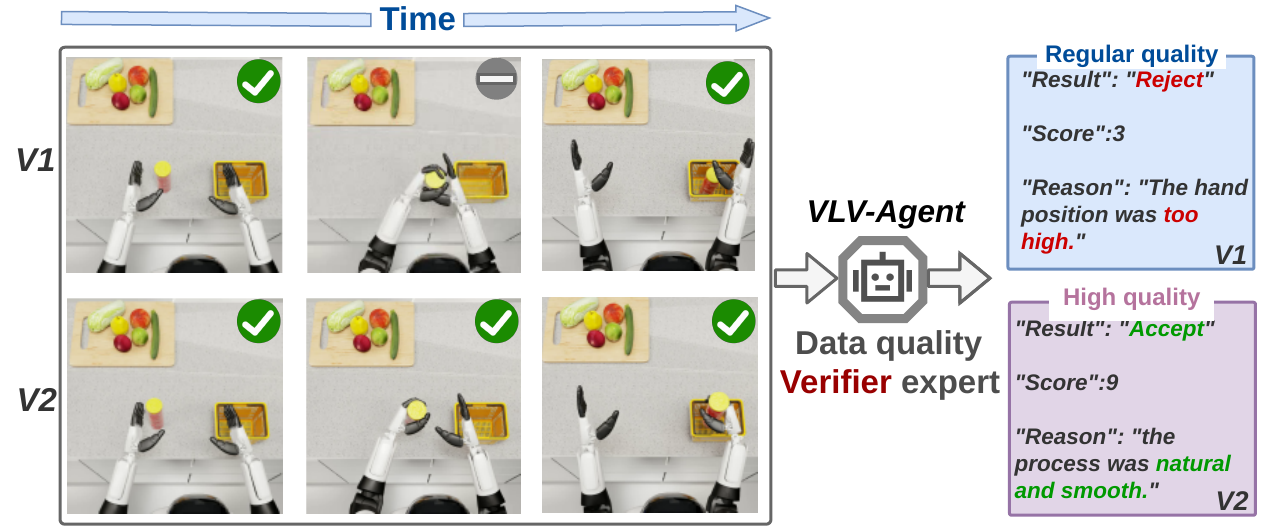}
    \caption{VLV analysis of successful trajectories with different quality levels. Top: regular-quality execution (moderate score). Bottom: high-quality execution (highest score).}
    \label{fig:3case}
    \vspace{-4mm}
\end{figure}

\subsection{The VLV as Quality Gatekeeper}
Beyond the rule-based Success Gate that ensures data validity by construction, we study the value of fine-grained quality gating among successful trajectories. As quantified in Fig. \ref{fig:collector_comparison}, the collector's success rate remains below 15\% in early iterations, and even among successful attempts, execution quality varies substantially. Comparing a baseline that retains all successful trajectories (\textbf{SuperTiny$^{-}$}) against our strategy that filters suboptimal successes (\textbf{SuperTiny}), Fig. \ref{fig:collector_comparison} reveals a significant performance gap. This ablation confirms that enforcing high trajectory quality—not just task completion—is essential for robust policy learning



Qualitative Visualization: As visualized in Fig. \ref{fig:3case}, the VLV effectively differentiates execution quality among successful trajectories: it assigns moderate scores to regular-quality executions (Top) and rewards high-quality, smooth completions with the highest score (Bottom). This demonstrates human-aligned scoring that guides self-evolution.

\section{conclusions}

In this paper, we presented Seed2Scale, a self-evolving data engine that addresses the data scarcity challenge in robot learning through a dual-stage paradigm combining self-evolution with parallel worlds expansion. By orchestrating a heterogeneous synergy among a lightweight collector, a pretrained verifier, and a target policy model, we show that the proposed pipeline can generate additional VLV-filtered trajectories from a small number of human demonstrations in the evaluated manipulation tasks. The Parallel Worlds expansion further decouples visual diversity from spatial exploration, facilitating transfer to unseen scenes. Experiments on the evaluated tasks show significant performance improvements across self-evolution iterations and successful deployment without real-world fine-tuning. In the future, we aim to extend Seed2Scale to long-horizon tasks and whole-body manipulation, while exploring tighter collector-verifier integration for more efficient and scalable self-evolution.


\bibliographystyle{IEEEtran}  
\bibliography{IEEEfull}     

@article{zhao2023learning,
  title={Learning fine-grained bimanual manipulation with low-cost hardware},
  author={Zhao, Tony Z and Kumar, Vikash and Levine, Sergey and Finn, Chelsea},
  journal={arXiv preprint arXiv:2304.13705},
  year={2023}
}

@article{kim2024openvla,
  title={Openvla: An open-source vision-language-action model},
  author={Kim, Moo Jin and Pertsch, Karl and Karamcheti, Siddharth and Xiao, Ted and Balakrishna, Ashwin and Nair, Suraj and Rafailov, Rafael and Foster, Ethan and Lam, Grace and Sanketi, Pannag and others},
  journal={arXiv preprint arXiv:2406.09246},
  year={2024}
}

@inproceedings{kim2025openvla,
  title={OpenVLA: An Open-Source Vision-Language-Action Model},
  author={Kim, Moo Jin and Pertsch, Karl and Karamcheti, Siddharth and Xiao, Ted and Balakrishna, Ashwin and Nair, Suraj and Rafailov, Rafael and Foster, Ethan P and Sanketi, Pannag R and Vuong, Quan and others},
  booktitle={Conference on Robot Learning},
  pages={2679--2713},
  year={2025},
  organization={PMLR}
}

@article{chi2023diffusion,
  title={Diffusion policy: Visuomotor policy learning via action diffusion},
  author={Chi, Cheng and Xu, Zhenjia and Feng, Siyuan and Cousineau, Eric and Du, Yilun and Burchfiel, Benjamin and Tedrake, Russ and Song, Shuran},
  journal={The International Journal of Robotics Research},
  pages={02783649241273668},
  year={2023},
  publisher={SAGE Publications Sage UK: London, England}
}

@misc{figure_helix,
  title = {Helix: A Vision-Language-Action Model for Generalist Humanoid Control},
  url = {https://www.figure.ai/news/helix},
  author = {{Figure AI}},
  year = {2025},
  month = {February},
}

@article{shukor2025smolvla,
  title={Smolvla: A vision-language-action model for affordable and efficient robotics},
  author={Shukor, Mustafa and Aubakirova, Dana and Capuano, Francesco and Kooijmans, Pepijn and Palma, Steven and Zouitine, Adil and Aractingi, Michel and Pascal, Caroline and Russi, Martino and Marafioti, Andres and others},
  journal={arXiv preprint arXiv:2506.01844},
  year={2025}
}

@article{intelligence2504pi0,
  title={$\pi$0.5: a vision-language-action model with open-world generalization, 2025},
  author={Intelligence, Physical and Black, Kevin and Brown, Noah and Darpinian, James and Dhabalia, Karan and Driess, Danny and Esmail, Adnan and Equi, Michael and Finn, Chelsea and Fusai, Niccolo and others},
  journal={URL https://arxiv. org/abs/2504.16054},
  year={2025}
}

@misc{nvidia2025gr00tn16,
  author       = {{NVIDIA GEAR Team}},
  title        = {{GR00T N1.6}: An Improved Open Foundation Model for Generalist Humanoid Robots},
  year         = {2025},
  month        = dec,
  howpublished = {NVIDIA Research},
  url          = {\url{https://research.nvidia.com/labs/gear/gr00t-n1_6/}},
}

@article{mandlekar2023mimicgen,
  title={Mimicgen: A data generation system for scalable robot learning using human demonstrations},
  author={Mandlekar, Ajay and Nasiriany, Soroush and Wen, Bowen and Akinola, Iretiayo and Narang, Yashraj and Fan, Linxi and Zhu, Yuke and Fox, Dieter},
  journal={arXiv preprint arXiv:2310.17596},
  year={2023}
}

@inproceedings{jiang2025dexmimicgen,
  title={Dexmimicgen: Automated data generation for bimanual dexterous manipulation via imitation learning},
  author={Jiang, Zhenyu and Xie, Yuqi and Lin, Kevin and Xu, Zhenjia and Wan, Weikang and Mandlekar, Ajay and Fan, Linxi Jim and Zhu, Yuke},
  booktitle={2025 IEEE International Conference on Robotics and Automation (ICRA)},
  pages={16923--16930},
  year={2025},
  organization={IEEE}
}

@article{pomponi2025dynamimicgen,
  title={DynaMimicGen: A Data Generation Framework for Robot Learning of Dynamic Tasks},
  author={Pomponi, Vincenzo and Franceschi, Paolo and Baraldo, Stefano and Roveda, Loris and Avram, Oliver and Gambardella, Luca Maria and Valente, Anna},
  journal={arXiv preprint arXiv:2511.16223},
  year={2025}
}

@inproceedings{ameperosa2025rocoda,
  title={Rocoda: Counterfactual data augmentation for data-efficient robot learning from demonstrations},
  author={Ameperosa, Ezra and Collins, Jeremy A and Jain, Mrinal and Garg, Animesh},
  booktitle={2025 IEEE International Conference on Robotics and Automation (ICRA)},
  pages={13250--13256},
  year={2025},
  organization={IEEE}
}

@article{jing2025humanoidgen,
  title={HumanoidGen: Data generation for bimanual dexterous manipulation via LLM reasoning},
  author={Jing, Zhi and Yang, Siyuan and Ao, Jicong and Xiao, Ting and Jiang, Yu-Gang and Bai, Chenjia},
  journal={arXiv preprint arXiv:2507.00833},
  year={2025}
}

@article{collaboration2023open,
  title={Open X-Embodiment: Robotic learning datasets and RT-X models},
  author={Collaboration, OX-Embodiment and O’Neill, Abby and Rehman, Abdul and Gupta, Abhinav and Maddukuri, Abhiram and Gupta, Abhishek and Padalkar, Abhishek and Lee, Abraham and Pooley, Acorn and Gupta, Agrim and others},
  journal={arXiv preprint arXiv:2310.08864},
  volume={1},
  number={2},
  year={2023}
}

@article{ye2024latent,
  title={Latent action pretraining from videos},
  author={Ye, Seonghyeon and Jang, Joel and Jeon, Byeongguk and Joo, Sejune and Yang, Jianwei and Peng, Baolin and Mandlekar, Ajay and Tan, Reuben and Chao, Yu-Wei and Lin, Bill Yuchen and others},
  journal={arXiv preprint arXiv:2410.11758},
  year={2024}
}

@article{wang2025unified,
  title={Unified vision-language-action model},
  author={Wang, Yuqi and Li, Xinghang and Wang, Wenxuan and Zhang, Junbo and Li, Yingyan and Chen, Yuntao and Wang, Xinlong and Zhang, Zhaoxiang},
  journal={arXiv preprint arXiv:2506.19850},
  year={2025}
}

@article{bahl2022human,
  title={Human-to-robot imitation in the wild},
  author={Bahl, Shikhar and Gupta, Abhinav and Pathak, Deepak},
  journal={arXiv preprint arXiv:2207.09450},
  year={2022}
}

@article{dessalene2025embodiswap,
  title={EmbodiSwap for Zero-Shot Robot Imitation Learning},
  author={Dessalene, Eadom and Mantripragada, Pavan and Maynord, Michael and Aloimonos, Yiannis},
  journal={arXiv preprint arXiv:2510.03706},
  year={2025}
}

@article{liu2025immimic,
  title={Immimic: Cross-domain imitation from human videos via mapping and interpolation},
  author={Liu, Yangcen and Shin, Woo Chul and Han, Yunhai and Chen, Zhenyang and Ravichandar, Harish and Xu, Danfei},
  journal={arXiv preprint arXiv:2509.10952},
  year={2025}
}

@article{hsieh2025dexman,
  title={DexMan: Learning Bimanual Dexterous Manipulation from Human and Generated Videos},
  author={Hsieh, Jhen and Tu, Kuan-Hsun and Hung, Kuo-Han and Ke, Tsung-Wei},
  journal={arXiv preprint arXiv:2510.08475},
  year={2025}
}

@article{zheng2026egoscale,
  title={EgoScale: Scaling Dexterous Manipulation with Diverse Egocentric Human Data},
  author={Zheng, Ruijie and Niu, Dantong and Xie, Yuqi and Wang, Jing and Xu, Mengda and Jiang, Yunfan and Casta{\~n}eda, Fernando and Hu, Fengyuan and Tan, You Liang and Fu, Letian and others},
  journal={arXiv preprint arXiv:2602.16710},
  year={2026}
}

@article{ye2026world,
  title={World Action Models are Zero-shot Policies},
  author={Ye, Seonghyeon and Ge, Yunhao and Zheng, Kaiyuan and Gao, Shenyuan and Yu, Sihyun and Kurian, George and Indupuru, Suneel and Tan, You Liang and Zhu, Chuning and Xiang, Jiannan and others},
  journal={arXiv preprint arXiv:2602.15922},
  year={2026}
}

@article{wu2023unleashing,
  title={Unleashing large-scale video generative pre-training for visual robot manipulation},
  author={Wu, Hongtao and Jing, Ya and Cheang, Chilam and Chen, Guangzeng and Xu, Jiafeng and Li, Xinghang and Liu, Minghuan and Li, Hang and Kong, Tao},
  journal={arXiv preprint arXiv:2312.13139},
  year={2023}
}

@article{hurst2024gpt,
  title={Gpt-4o system card},
  author={Hurst, Aaron and Lerer, Adam and Goucher, Adam P and Perelman, Adam and Ramesh, Aditya and Clark, Aidan and Ostrow, AJ and Welihinda, Akila and Hayes, Alan and Radford, Alec and others},
  journal={arXiv preprint arXiv:2410.21276},
  year={2024}
}

@article{bai2025qwen3,
  title={Qwen3-vl technical report},
  author={Bai, Shuai and Cai, Yuxuan and Chen, Ruizhe and Chen, Keqin and Chen, Xionghui and Cheng, Zesen and Deng, Lianghao and Ding, Wei and Gao, Chang and Ge, Chunjiang and others},
  journal={arXiv preprint arXiv:2511.21631},
  year={2025}
}

@article{yang2025qwen3,
  title={Qwen3 technical report},
  author={Yang, An and Li, Anfeng and Yang, Baosong and Zhang, Beichen and Hui, Binyuan and Zheng, Bo and Yu, Bowen and Gao, Chang and Huang, Chengen and Lv, Chenxu and others},
  journal={arXiv preprint arXiv:2505.09388},
  year={2025}
}

@article{tai2025realmirror,
  title={RealMirror: A Comprehensive, Open-Source Vision-Language-Action Platform for Embodied AI},
  author={Tai, Cong and Zheng, Zhaoyu and Long, Haixu and Wu, Hansheng and Xiang, Haodong and Long, Zhengbin and Xiong, Jun and Shi, Rong and Zhang, Shizhuang and Qiu, Gang and others},
  journal={arXiv preprint arXiv:2509.14687},
  year={2025}
}

@inproceedings{chen2024internvl,
  title={Internvl: Scaling up vision foundation models and aligning for generic visual-linguistic tasks},
  author={Chen, Zhe and Wu, Jiannan and Wang, Wenhai and Su, Weijie and Chen, Guo and Xing, Sen and Zhong, Muyan and Zhang, Qinglong and Zhu, Xizhou and Lu, Lewei and others},
  booktitle={Proceedings of the IEEE/CVF conference on computer vision and pattern recognition},
  pages={24185--24198},
  year={2024}
}

@article{zheng2025demonstrating,
  title={Demonstrating DVS: Dynamic virtual-real simulation platform for mobile robotic tasks},
  author={Zheng, Zijie and Li, Zeshun and Wang, Yunpeng and Xie, Qinghongbing and Zeng, Long},
  journal={arXiv preprint arXiv:2504.18944},
  year={2025}
}

@inproceedings{zhao2026real2edit2real,
  title={Real2edit2real: Generating robotic demonstrations via a 3d control interface},
  author={Zhao, Yujie and Fan, Hongwei and Chen, Di and Chen, Shengcong and Chen, Liliang and Li, Xiaoqi and Ren, Guanghui and Dong, Hao},
  booktitle={Proceedings of the IEEE/CVF Conference on Computer Vision and Pattern Recognition},
  pages={23106--23116},
  year={2026}
}

@article{sedlacek2026realm,
  title={Realm: A real-to-sim validated benchmark for generalization in robotic manipulation},
  author={Sedlacek, Martin and Yefanov, Pavlo and Ponimatkin, Georgy and Bardhan, Jai and Pilc, Simon and Fourmy, Mederic and Kazakos, Evangelos and Snoek, Cees GM and Sivic, Josef and Petrik, Vladimir},
  journal={IEEE Robotics and Automation Letters},
  year={2026},
  publisher={IEEE}
}

@inproceedings{heppert2024ditto,
  title={Ditto: Demonstration imitation by trajectory transformation},
  author={Heppert, Nick and Argus, Max and Welschehold, Tim and Brox, Thomas and Valada, Abhinav},
  booktitle={2024 IEEE/RSJ international conference on intelligent robots and systems (IROS)},
  pages={7565--7572},
  year={2024},
  organization={IEEE}
}

\end{document}